\pdfoutput=1

\documentclass[11pt]{article}

\usepackage[]{ACL2023}

\usepackage{times}
\usepackage{latexsym}
\usepackage[linguistics]{forest}
\usepackage{acronym}
\usepackage{multirow}
\usepackage{enumitem}
\usepackage{amssymb}

\usepackage[T1]{fontenc}

\usepackage[utf8]{inputenc}

\usepackage{microtype}

\usepackage{inconsolata}

\usepackage{amsmath}

\DeclareMathOperator*{\minimize}{minimize}
\title{A Quantitative Approach to Understand Self-Supervised Models\\as Cross-lingual Feature Extractors}

\author{
Shuyue Stella Li$^{1\star}$, Beining Xu$^{2\star}$, Xiangyu Zhang$^{1\star}$, Hexin Liu$^3$, Wenhan Chao$^2$, Leibny Paola Garcia$^1$ \\
  $^1$Center for Language and Speech Processing, Johns Hopkins University\\
  $^2$School of Computer Science and Engineering, Beihang University\\
  $^3$School of Electrical and Electronic Engineering, Nanyang Technological University\\
  \texttt{sli136, xzhan233, lgarci27@jhu.edu}
}

\begin{document}
\maketitle
\newacro{asr}[ASR]{automatic speech recognition}
\newacro{psr}[PSR]{Phonology-Syntax Ratio}
\newacro{ssl}[SSL]{Self-Supervised Learning}
\newacro{cca}[CCA]{Canonical Correlation Analysis}
\newacro{ldn}[LDN]{Normalized Levenshtein Distance}
\newacro{ldnd}[LDND]{modified normalized Levenshtein Distance}
\newacro{dnn}[DNN]{Deep neural network}
\begin{abstract}
In this work, we study the features extracted by English self-supervised learning (SSL) models in cross-lingual contexts and propose a new metric to predict the quality of feature representations. Using \ac{asr} as a downstream task, we analyze the effect of model size, training objectives, and model architecture on the models' performance as a feature extractor for a set of topologically diverse corpora. 
We develop a novel metric, the Phonetic-Syntax Ratio (PSR), to measure the phonetic and synthetic information in the extracted representations using deep generalized canonical correlation analysis. Results show the contrastive loss in the wav2vec2.0 objective facilitates more effective cross-lingual feature extraction. 
There is a positive correlation between PSR scores and ASR performance, suggesting that phonetic information extracted by monolingual SSL models can be used for downstream tasks in cross-lingual settings. The proposed metric is an effective indicator of the quality of the representations and can be useful for model selection.\footnote{We make our work open-source for further explorations: \url{https://github.com/stellali7/SSL_PSR}}
\end{abstract}

\section{Introduction}
\begin{figure}
    \centering
    \includegraphics[width=0.48\textwidth]{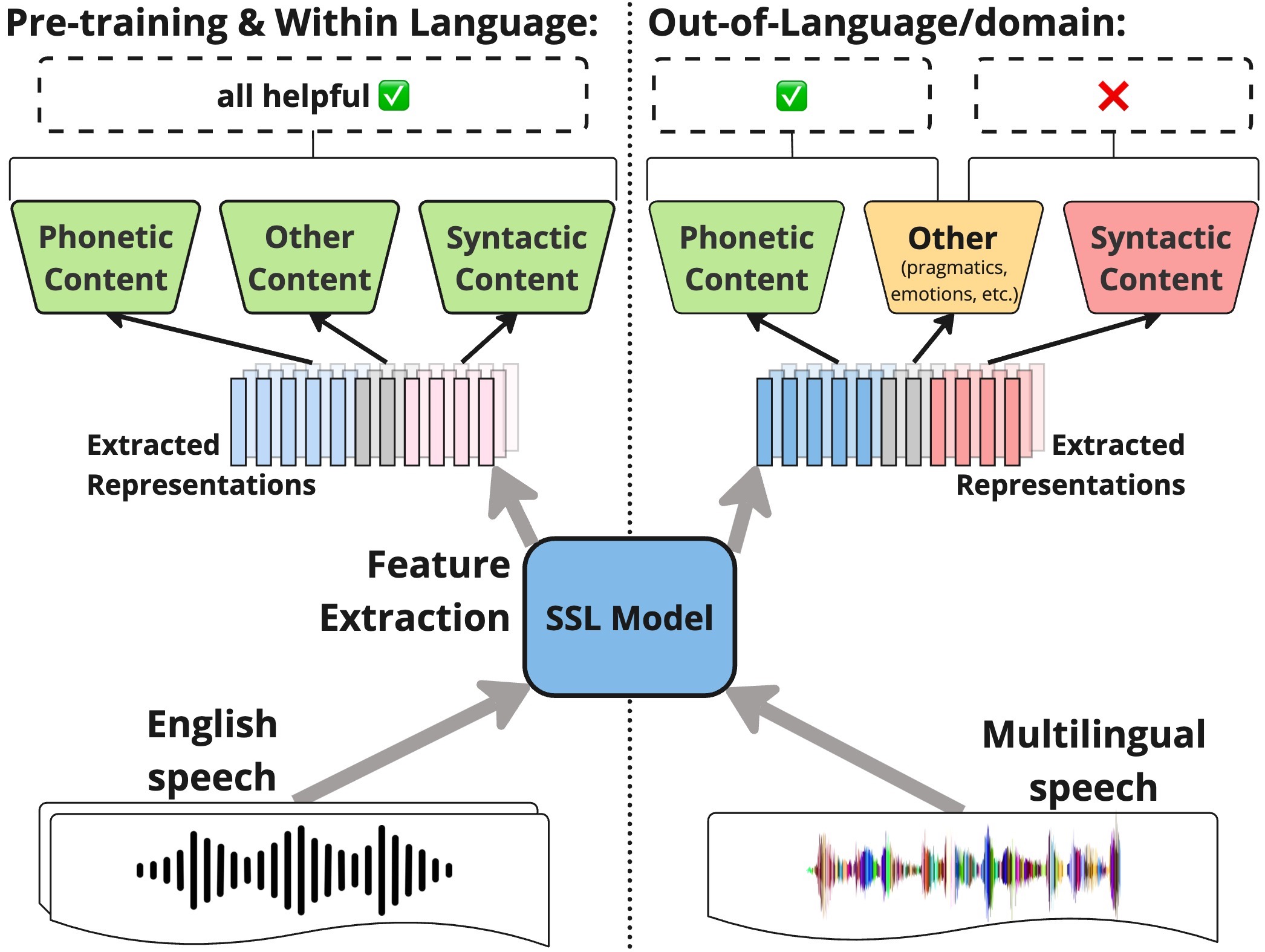}
    \small{\caption{Speech data of English (in-domain) and other languages (out-of-domain) are passed through the SSL models to extract speech representations. All information is expected to aid downstream tasks in English while phonetic content is expected to be useful for out-of-domain downstream tasks; ``other" content may include speaker information, etc.}\label{fig:first}}\vspace{-4mm}
\end{figure}

\textbf{\ac{ssl}} has become a paradigm for learning feature representations from unlabeled data \cite{liu2021self}. In speech processing, self-supervised approaches for learning speech representation are often used to extract features for downstream tasks. These representations can replace the handcrafted feature such as Mel Spectrum or MFCC in many tasks as they are able to extract high-level properties in the speech data \cite{mohamed2022self, chung2019unsupervised}. 



\textbf{English \ac{ssl} Models} take advantage of the high availability of English data and outperform traditional feature extraction methods on a range of downstream tasks in English \cite{chen2022wavlm, hsu2021hubert, liu2020mockingjay}. 
Since the acoustic and phonetic information of human speakers across languages share a level of similarity, it is crucial to study the cross-lingual transfer performance of English \ac{ssl} models as a feature extractor for non-English audio data \cite{li2020improving, cho2018multilingual}. This will enhance our understanding of the composition of knowledge learned during pre-training, allowing more efficient use of data during model selection. Furthermore, if we are able to use English monolingual models effectively in multilingual downstream tasks, the high cost of training massive multilingual speech models such as XLSR \cite{babu2021xls, conneau2020unsupervised} and mSLAM \cite{bapna2022mslam} can be reduced by explicitly incorporating architectural designs promoting cross-lingual transfer.
Therefore, the first purpose of this paper is to investigate the factors that improve the ability of monolingual \ac{ssl} models to extract useful speech representations for \ac{asr} tasks in typologically diverse languages.

The second objective of our study is to analyze the amount of phonetic information versus syntactic information learned by the model during training, and how the phonetic-syntax composition in the model impacts the extracted features. Phonetic content directly impacts the learned phonological structure in the representations. Explicit integration of phonological knowledge has proven to be extremely successful in speech processing \cite{zhan2021self}. On the other hand, semantic and syntactic knowledge learning in the target language during fine-tuning is needed for \ac{asr} tasks so that the \ac{ssl} models do not retain source language semantics and syntax, implying syntactic information might be harmful for cross-lingual feature extraction \cite{li2020improving}. 

As shown in Figure \ref{fig:first}, we expect the pre-trained \ac{ssl} models to efficiently extract phonetic, syntactic, and other contents to help downstream tasks in English \cite{chung2021similarity}. At the same time, the extracted phonetic information in out-of-domain and multilingual situations should also aid downstream performance. Therefore, we propose a novel metric to quantify the amount of helpful phonetic information.
To the best of our knowledge, this study is the first to quantitatively understand the capabilities and limits of \ac{ssl} models from a linguistic perspective. Our contributions include:
\begin{itemize}[noitemsep,topsep=0pt,leftmargin=10pt]
    \item We examine five \ac{ssl} models with different sizes, data preparation methods, and training objectives by analyzing their cross-lingual generalizability as feature extractors on the \ac{asr} task.
    \item We propose a new metric, \ac{psr}, to measure the phonetic and syntactic content extracted by an \ac{ssl} model on any given out-of-domain/language dataset. A higher \ac{psr} score correlates to a better \ac{asr} performance.
    \item We localize the phonetic content in the \ac{ssl} model to specific layers using the trained layer-wise weights for the feature representations.
\end{itemize}

\section{Related Work}
\subsection{Self-Supervised Models}
Self-supervised learning (SSL) \cite{liu2021self, bengio2013representation, raina2007self} takes advantage of easily accessible unlabeled data to learn a model and then produces universal representations by solving upstream tasks \cite{liu2022audio}. Then, the pre-trained \ac{ssl} model can be used to process unseen data based on its previous knowledge and handle multiple downstream tasks. \ac{ssl} models have achieved superior performance in natural language processing \cite{devlin2018bert, MatthewEPeters2018DeepCW}, computer vision \cite{chen2020simple, IshanMisra2020SelfSupervisedLO}, speech processing \cite{chung2016audio, chi2021audio}, and especially \ac{asr} \cite{baevski2020effectiveness, ravanelli2020multi, jiang2021further}. In our work, we study a number of SSL models and their feature extraction ability when presented with input from other languages.

\subsection{Audio Feature Extraction}
Before any downstream speech processing tasks, the audio data is converted to high-dimensional feature vectors through an audio feature extraction system \cite{moffat2015evaluation}. Classic methods, such as Mel-Frequency Cepstral Coefficients (MFCCs), Linear Predictive Coding (LPC), and Perceptual Linear Prediction (PLP) extract cepstral coefficients that contain low-level acoustic features \cite{dave2013feature, shanthi2013review}. Researchers have also delved into neural-based models, leveraging pre-trained models on large-scale datasets to boost performance \cite{chi2021audio}. While progress has been remarkable, challenges such as robustness to noise variations and interpretability of learned features continue to stimulate further research in this domain \cite{mohamed2022self}. In our work, we explore the robustness of the monolingual \ac{ssl} models when generalized to multilingual settings, from which we interpret the features extracted by these models. 

\begin{figure*}[ht]
    \centering
    \includegraphics[width=0.79\textwidth]{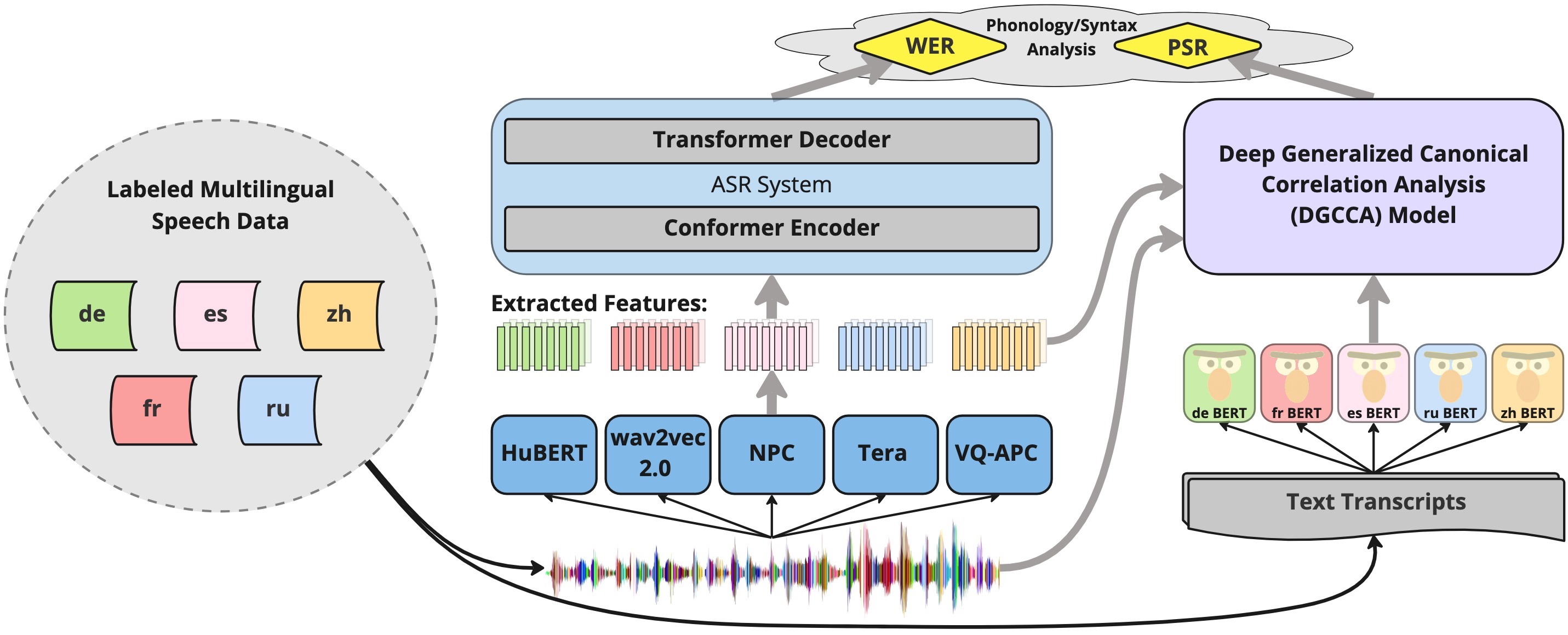}\vspace{-1mm}
    \small{\caption{The pipeline to measure the performance of \ac{ssl} model on different languages. We first use each \ac{ssl} model as a feature extractor for data in each language and compute a WER score for the \ac{asr} task. Then, we calculate the \ac{psr} of the representations to analyze the correlation between the \ac{asr} performance and the \ac{psr} score.}\label{fig:pipeline}}\vspace{-3mm}
\end{figure*}

\subsection{Automatic Speech Recognition (ASR)} 
\ac{asr} transcribes given audio to text in the script of the spoken language \cite{malik2021automatic, yu2016automatic}. \ac{dnn} based techniques \cite{hinton2012deep} have boosted the accuracy of \ac{asr} by replacing the traditional Gaussian Mixture Model in cascaded systems involving separate acoustic, language, and lexicon components \cite{li2022recent}. End-to-end models \cite{graves2014towards, chorowski2014end, bahdanau2016end, collobert2016wav2letter} have recently become a breakthrough in the speech community, directly translating an input speech sequence into an output text sequence with a single model. Some publicly available and commonly used toolkits include Kaldi \cite{povey2011kaldi}, CMU Sphinx \cite{45616}, SpeechBrain \cite{ravanelli2021speechbrain} and ESPNet \cite{watanabe2018espnet}.

\subsection{Analysis Methods of SSL Models}
There has been extensive research on analyzing supervised speech models \cite{belinkov2019analysis, palaskar2019learned, prasad2020accents}. However, research on \ac{ssl} models, especially in the speech domain, is still relevantly under-explored. Some recent work in this field includes a similarity analysis of self-supervised speech representations, in which they only looked into simpler models such as APA, CPC, and MPC \cite{chung2021similarity}. 
\citet{liu2022efficient} attempted to distinguish useful representations in \ac{ssl} models for spoken language identification and reduce spurious information in the representations, but was limited to a specific task. 
\citet{pasad2021layer} and \citet{pasad2023comparative} analyzed the layer-wise acoustic-linguistic content of pre-trained models by performing layer-independent Canonical Correlation Analysis (CCA) \cite{hardoon2004canonical} on English data. However, since the features extracted by \ac{dnn} models often have high dimensionality \cite{georgiou2020survey}, CCA is limited in its ability to freely model complex nonlinear relationships.

\subsection{Cross-lingual Knowledge Transfer}
Cross-lingual transfer learning has gained attention in the field as it effectively mitigates resource constraints and language-specific challenges, but most importantly to our work, it requires the model to be able to adapt to unseen situations such as a new language \cite{khurana2023improved, conneau2019unsupervised}.
Effective cross-lingual transfer for speech processing requires the model to have a high-level understanding of both text linguistics and phonetics. 
Previous work has shown that multilingual models generalize well to target languages \cite{conneau2020unsupervised, singh2019xlda, radford2023robust}. \citet{lauscher2020zero} shows that the quality of the cross-lingual transfer is correlated with the linguistic similarity between the source and target languages.
Inspired by this, we use English monolingual models in our work to better compare the linguistic distance between the pre-train data and the target data. 
Studying the generalization ability of monolingual models to unseen languages allows us to better analyze the learned representations and localize the factors that facilitate cross-lingual transfer for more efficient model design.

\section{Analysis Methods}
As shown in Figure \ref{fig:pipeline}, we first use the \ac{ssl} models trained on English to extract speech representations on audio data from German (de), French (fr), Spanish (es), Russian (ru), and Chinese (zh). Then, we use the \ac{asr} task to evaluate the quality of the extracted features against a Mel Spectrum baseline in Section \ref{sec:method:multilingual}. We correlate the WER scores to traditional measures of linguistic distance in Section \ref{sec:method:ling}. Finally, we quantitatively evaluate the phonetic and syntactic content in the extracted features for each language, as described in Section \ref{sec:method:psr}.

\subsection{Measuring Multilingual Generalizability}\label{sec:method:multilingual}
We use the standard \ac{asr} task on 5 genealogically and typographically diverse languages to evaluate the generalizability of the English \ac{ssl} models as a cross-lingual feature extractor. To fairly compare the models, we freeze the parameters of the models and use the same downstream architecture (Conformer + Transformer) for all \ac{ssl} models and the Mel Spectrum baseline feature extractor. We also use the same language model setup and beam size during decoding.

Our pipeline is shown in Figure \ref{fig:pipeline}. We select \ac{ssl} models based on their training methods. These upstream \ac{ssl} models can be categorized into \textbf{masked reconstruction model}: Tera~\cite{liu2021tera} and NPC~\cite{liu2020non}; \textbf{masked prediction model}: HuBERT~\cite{hsu2021hubert}; \textbf{auto-regressive reconstruction model}: VQ-APC~\cite{chung2020vqapc}; and \textbf{contrastive model}: wav2vec2.0~\cite{baevski2020wav2vec}. Inspired by the setup in SUPERB \cite{yang2021superb} and ELMO \cite{peters-etal-2018-deep}, we take the weighted sum from all layers as the extracted representation, and the weight vector is updated during training. 

For the downstream model, we use the Conformer \cite{gulati2020conformer} as the encoder and the Transformer \cite{vaswani2017attention}, which has achieved state-of-the-art (SOTA) results in many speech recognition tasks \cite{ma2021end}. 
During data analysis, we isolate the effect of the \ac{ssl} model as a feature extractor by taking the difference ($\Delta$) between the \ac{ssl} feature extractor and the Mel Spectrum baseline performance. This eliminates any potential noise introduced by data size differences, speech formality levels, and other linguistic differences between languages, allowing a fair comparison between different \ac{ssl} models. When decoding, we use a simple RNN as a language model and keep the parameters consistent across all tasks.

\begin{figure}[ht]
    \begin{center}\vspace{-2.8mm}
    \scalebox{0.95}{
    \hspace{-0.6cm}\begin{forest}
      [
        [Indo-European
         [Germanic
           [\textbf{de}]
           [\textbf{\textcolor{red}{en}}]
         ]
         [Romance
           [\textbf{fr}]
           [\textbf{es}]
         ]
         [Slavic
           [\textbf{ru}]
         ]
        ]
        [Sino-Tibetan
          [\textbf{zh}]
        ]
      ]
    \end{forest}}
    \end{center}\vspace{-4.5mm}
    \caption{Phylogenetic Tree of Target Langauges}\vspace{-3.5mm}
    \label{fig:phylotree}
\end{figure}
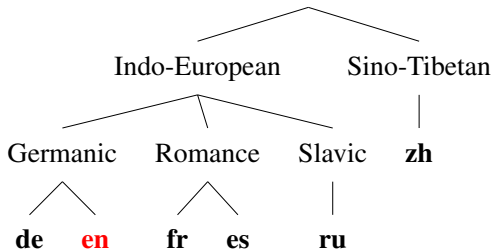

\subsection{Measuring Linguistic Distance} \label{sec:method:ling}
We examine the performance of self-supervised models on languages across a diverse range of families and groups in order to investigate the relationship between model performance and linguistic distance. In our analysis, we employ the phylogenetic tree in Figure \ref{fig:phylotree} derived from the theory of language evolution with genetic distance equaling the Levenshtein distance \cite{serva2008indo} as a measure of linguistic distance. Since languages evolve with both their written and spoken forms, the phylogenetic tree will contain the most comprehensive information about the language.

\subsection{Measuring Phonetic \& Syntactic Content} \label{sec:method:psr}
In this section, we describe approaches to quantify phonetic and syntactic content in the extracted speech representations of \ac{ssl} models.

\begin{figure}[h]
    \centering
    \includegraphics[width=0.4\textwidth]{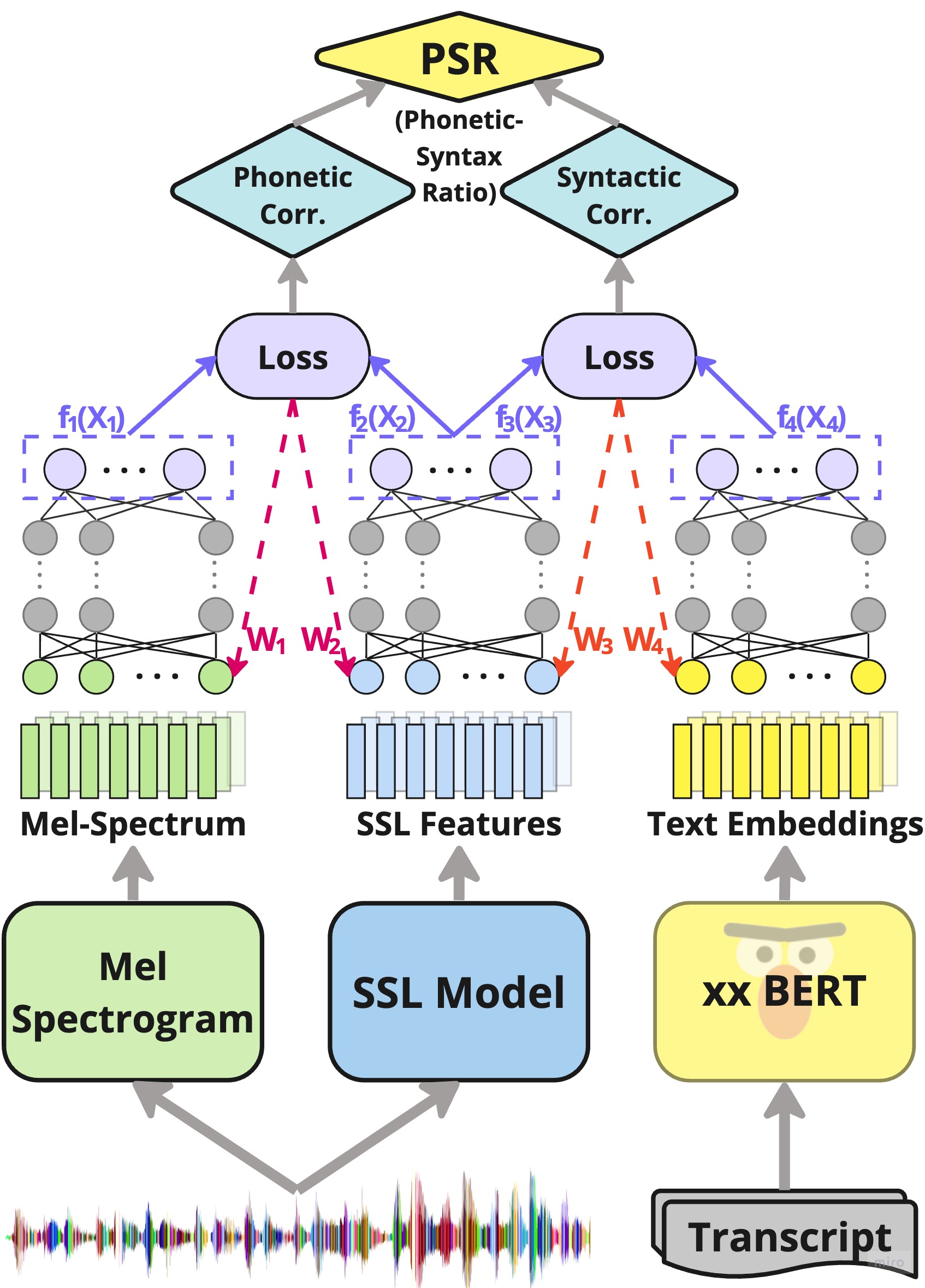}\vspace{-1.5mm}
    \small{\caption{DGCCA pipeline. The model aims to compare the representation extracted by the \ac{ssl} model to the pure acoustic representation (from Mel Spectrum) and pure syntactic/semantic representation (from BERT).} \label{fig:dgcca}}\vspace{-3.5mm}
\end{figure}

\subsubsection{DGCCA} \label{sec:dgcca}
In order to better analyze the phonetic and syntactic content of the features, we use a tool called Deep Generalized Canonical Correlation Analysis (DGCCA), which is a deep learning technique that measures the nonlinear relationship between arbitrarily many views of the data and learns a view-independent representation \cite{benton2017deep}. DGCCA effectively quantifies the phonetic and syntactic content of \ac{ssl} models when treating the features extracted with different models as different views of the same data.

As shown in Figure \ref{fig:dgcca}, DGCCA takes $N$ pairs of data vectors across $J$ views as input and returns a correlation score as a measure of the similarity between the vectors. Using standard back-propagation to optimize the weight matrices $W_j=\{W_1^j,\ldots,W_{K_j}^j\}$, we try to find the linear transformation $U_j\in\mathbb{R}^{d_j\times N}$ of $f_i(X_j)\in\mathbb{R}^{o_j}$ constrained by $GG^T = I_r$ such that:
\vspace{-1.3mm}\begin{equation}
    \minimize_{U_j\in\mathbb{R}^{d_j\times N}, G\in\mathbb{R}^{r\times N}}  \sum_{j=1}^J \|G - U_j^T f_j(X_j) \|^2_F\,,\vspace{-0.8mm}
\end{equation}
where $X_j\in\mathbb{R}^{d_j\times N}$ is the input feature vectors of the $j^{th}$ view; $f_j$ is the function learned using a multilayer perceptron of $K_j$ layers; $d_j$ is the dimension of the $j^{th}$ view and $r$ is the dimension of the learned representation $G$. 

In our case, $N$ is the number of utterances in the test data, where we have the \ac{ssl} features and Mel Spectrum features of each utterance, as well as the BERT representations of its transcript. The monolingual BERT model in each target language is used when extracting the textual representations.

Features extracted by the \ac{ssl} models, pure phonetic features (Mel Spectrum), and pure textual features (BERT representations) can be considered as different views ($f_i$) of the data. The correlation scores between different views are the loss of the converged DGCCA network. We compute the correlation scores between each of the latter two views and the \ac{ssl} features. The correlation scores between the \ac{ssl} features and the Mel Spectrum measure the \textbf{Phonetic Content} in the extracted features; the correlation scores between the \ac{ssl} features and the BERT representations measure the \textbf{Syntactic Content} in the extracted features.

\subsubsection{Phonetic-Syntax Ratio (PSR)}
We introduce a new metric: the Phonetic-Syntax Ratio (PSR) in order to quantitatively investigate the phonetic and syntactic content on \ac{ssl} representation. As described in Section \ref{sec:dgcca}, the similarity to phonetic features and the similarity to syntactic features of the SSL representations are both optimized and quantified as correlation scores when training the DGCCA network. We define the PSR as the ratio between the phonetic correlation score and syntactic correlation score, weighted equally among all data points:\\
\vspace{-1mm}\begin{equation}\hspace{-1mm}\label{eq:psr}
    PSR = (\frac{1}{n}\sum_{i=1}^n\frac{phonetic\ score_i}{syntax\ score_i}-1)\cdot100\%,
\end{equation}
where phonetic scores and syntax scores are the output of DGCCA when the \ac{ssl} representations are fed in with the Mel Spectrum and BERT contextualized embedding, respectively. The PSR score is model-agnostic and language-agnostic, and can be used for a range of contrastive analysis for inferring cross-lingual transferability. 

\section{Experimental Setup}

\subsection{Datasets}
We investigate the cross-lingual adaptation capability of English \ac{ssl} models in five languages. For training the \ac{asr} models, we use the Mozilla Common Voice 5.1 dataset \cite{ardila2019common} for German, French, Spanish, and Russian, and we use the OpenSLR ST-CMDS-20170001\_1 Free ST Chinese Mandarin Corpus\footnote{http://www.openslr.org/38} for Chinese. The Common Voice English test set is used for DGCCA analysis. More details about the datasets are in Table \ref{tab:dataset}.

\begin{table}[h!]\centering
\scalebox{0.93}{\begin{tabular}{c|ccccc}
\hline
Lang & hr & voices & train & dev & test  \\
\hline
de   & 751 & 11,731 & 196,464 & 15,341 & 15,341 \\
fr   & 605 & 11,960 & 254,863 & 15,621 & 15,621 \\
es   & 522 & 18,906 & 138,878 & 14,860 & 14,860 \\
ru   & 117 & 927 & 13,189 & 7,242 & 7,307 \\
zh   & - & - & 92,280 & 4,299 & 4,483 \\
en   & 1933 & 61528 & 435,947 & 16,029 & 16,029 \\
\hline
\end{tabular}}\vspace{-2mm}
\small{\caption{\label{tab:dataset} Dataset description; the number of hours, voices, and utterances for each split. Hour and voice statistics for the Chinese corpus are not available as it is distributed after preprocessing. The number of speakers for the Chinese dataset is 855. Train and dev splits of English were not used.}}\vspace{-3mm}
\end{table}

\begin{table*}[ht]
\centering
\scalebox{0.94}{\begin{tabular}{ccccccc}
\hline
\textbf{Model} & \textbf{architecture} & \textbf{train objective} & \textbf{model size} & \textbf{pre-train} & \textbf{input} & \textbf{stride}\\
\hline
 HuBERT-\textsc{Base}   & \multirow{2}{*}{CNN + Transformer} & \multirow{2}{*}{Predictive}  & 95m & LS-960 & \multirow{2}{*}{wav} & \multirow{2}{*}{20ms}\\
 HuBERT-\textsc{Large}   & & & 317m & LL-60k & &\\
\hline
 wav2vec2-\textsc{Base} & \multirow{2}{*}{CNN + Transformer} & Contrastive & 95m & LS-960 & \multirow{2}{*}{wav} &\multirow{2}{*}{20ms}\\
 wav2vec2-\textsc{Large} & & \hspace{4mm} + Diversity & 317m & LV-53.2k & &\\
\hline
 NPC        & Masked Conv Block & L1 Reconstruction & 19.4m & LS-C-360 & Mel &10ms\\
\hline
\multirow{2}{*}{TERA-\textsc{Base}} & Unidirectional LSTM & \multirow{2}{*}{L1 Reconstruction} & \multirow{2}{*}{21.3m} & \multirow{2}{*}{LS-C-100} & \multirow{2}{*}{Mel} & \multirow{2}{*}{10ms}\\
            & + Prediction Network &  &  &  & \\
\hline
 VQ-APC     & Unidirectional LSTM & L1 Reconstruction & 4.63m & LS-C-360 & Mel &10ms\\
\hline
\end{tabular}} \vspace{-2.5mm}
\small{\caption{\label{tab:models} \ac{ssl} Model Summary. For the pre-training data description, LS = Librispeech, LS-C = Librispeech-clean, LL = Libri-light, and LV = Libri-vox.}}\vspace{-3mm}
\end{table*} 

\subsection{Multilingual Generalizability Setup}
We use the \ac{asr} performance on a range of typologically diverse languages as a metric to infer the models' multilingual generalizability. In order to fairly compare the performance of each \ac{ssl} model in different language datasets, we use the same downstream model for all languages and features and focus on the within-language difference between the \ac{ssl} model and the baseline model. 


\paragraph{Self-supervised feature extractors} We examine a number of English SSL speech models including HuBERT \cite{hsu2021hubert}, wav2vec 2.0 \cite{collobert2016wav2letter}, NPC \cite{liu2020non}, TERA \cite{liu2021tera}, and VQ-APC \cite{chung2020vqapc} with model details shown in Table \ref{tab:models}. 
Unlike the baseline model, we use a smaller learning rate considering that self-supervised training usually uses a small learning rate. We use a learning rate of 0.0025 with 40000 warmup steps. 

\paragraph{Model architectures}
After multilingual features are extracted, we use a standard Conformer encoder and a Transformer decoder in our downstream ASR model and a stacked RNN as the language model during decoding. More details on the ASR model architecture and training are in Appendix \ref{app:asr_arch}.

\subsection{\ac{psr} Computation}
We use the DGCCA pipeline shown in Figure \ref{fig:dgcca} to compute the PSR scores for each langauge. The DGCCA model used consists of an MLP network with a Linear layer, a Sigmoid function, and a Batch Norm layer. Each group of tensors has one MLP network, and its output is passed into the DGCCA loss. We used SGD to optimize the network with a learning rate of 1e-6.
We use features extracted by the HuBERT model from five different languages (German, French, Spanish, Russian, and English) and also extract its corresponding Mel Spectrum and BERT features. Chinese PSR is not reported because CER was used to evaluate the ASR performance, hence the comparison across languages would not be fair (more details in Section \ref{sec:results:linguistic}). When calculating the correlation scores, we use the test set in each target language as input to the DGCCA model with a batch size of 32. Details on the implementation and hardware of the SSL models and the DGCCA model can be found in Appendix \ref{app:hardware}.

\begin{table*}[ht]
\centering
\scalebox{0.87}{\begin{tabular}{c|cc|cc|cc|cc|cc|cc}
\hline
Model/Lang & de &$\Delta$ &fr &$\Delta$    &es &$\Delta$      &ru   &$\Delta$  &zh  &$\Delta$     &Avg. & $\Delta$\\
\hline
 Mel (Baseline)        &10.0   & - &\textbf{15.8}&- &\textbf{11.5}&-  &7.9  &-        &9.4 &-           &\textbf{10.92}   &-\\
\hline
 HuBERT-\textsc{Base}   &11.3  &1.3   &16.5  &\textbf{0.7}&13.1 &1.6  &7.8  &-0.1    &9.8 &0.4          &11.70   &0.78\\
 HuBERT-\textsc{Large}  &12.4  &2.4   &16.6  &0.8         &12.0 &\textbf{0.5}&8.3&0.4&\textbf{9.1}&\textbf{-0.3} &11.68&0.76\\
 wav2vec2-\textsc{Base} &11.8  &1.8   &16.7  &0.9         &13.4 &1.9  &8.5  &0.6     &9.8 &0.4          &12.04 &1.12\\
 wav2vec2-\textsc{Large}&\textbf{9.2} &\textbf{-0.8}      &16.6 &0.8  &12.3 &0.8 &\textbf{7.6}&\textbf{-0.3} &9.4 & 0  & 11.04 &\textbf{0.10}\\
 NPC                    &16.2  &6.2   &18.1  &2.3         &16.1 &4.6  &11.0 &3.1     &10.7 &1.3         &14.42 &3.5\\
 TERA-\textsc{Base}     &15.6  &5.6   &17.1  &1.3         &14.8 &3.3  &10.3 &2.4     &10.0 &0.6         &13.56  &2.64\\
 VQ-APC       &13.5   &3.5  &17.2 &1.4  &17.3 &5.8     &12.1 &4.2         &10.8 & 1.4  &14.18 &3.26\\
\hline
Avg.        &12.86&2.86 &16.97 &1.17   &14.14&2.64      &9.37 &1.47    &9.94 &0.54       &-  &-\\
\hline
\end{tabular}}\vspace{-2mm}
\small{\caption{\label{tab:asrresults} Word Error Rate (WER) of German (de), French (fr), Spanish (es), and Russian (ru). For Chinese (zh), we apply Character Error Rate (CER) as the evaluation metric. $\Delta$ is the difference from Baseline, the lower the better. wav2vec2.0-\textsc{Large} achieves the best performance and the Transformer-based models generally perform better.}} \vspace{-4mm}
\end{table*}

\section{Results and Analysis}\vspace{-1mm}
\subsection{Multilingual Generalizability}\label{sec:results:model}\vspace{-0.5mm}
Results from the multilingual \ac{asr} tasks are shown in Table \ref{tab:asrresults}, with both WER scores and the difference from the Mel Spectrum baseline ($\Delta$).

In the zero-shot setting, it is generally expected that the \ac{ssl} feature extractor trained on English, without any domain adaptation, performs poorly on the cross-lingual \ac{asr} tasks compared to the Mel spectrum baseline. Although it can extract higher-dimensional features, additional English syntactic information in the \ac{ssl} model can be projected onto the new language \cite{georgiou2020survey}. Therefore, the purpose of this experiment is not to improve the SOTA results but rather to probe the \ac{ssl} models for further phonetic-syntactic analysis. 

There are five \ac{ssl} models being evaluated in this experiment in five languages. The column \verb+Avg+ on the right marginal of Table \ref{tab:asrresults} shows the overall performance of each \ac{ssl} model in all languages. In general, wav2vec2.0-\textsc{Large} significantly outperforms other feature extractors and has a consistent result across languages. There are two instances in which wav2vec2.0-\textsc{Large} outperforms the pure acoustic Mel Spectrum baseline. This can be attributed to the cross-lingual phonetic information transfer that the model learned from English pre-training. 

\subsubsection{Effect of Training Objectives} \label{sec:results:model:training}
The HuBERT and wav2vec2.0 models consistently perform better than NPC, TERA, and VQ-APC. HuBERT and wav2vec2.0 both effectively combine CNN encoders with Transformers in their architecture. The attention mechanism allows the models to effectively encode speech features into the latent embedding space and learn contextualized representations. Both HuBERT and wav2vec2.0 use similar architectures and identical pre-training data and setups.
However, HuBERT as a cross-lingual feature extractor does not perform as well due to its predictive loss compared to the \textbf{contrastive loss} of wav2vec2.0. The masked prediction task during HuBERT pre-training forces the model to learn the language model as well as the acoustic model from continuous English speech inputs \cite{hsu2021hubert}, so the model might be overfitted to English syntax.

Now we discuss the performance of NPC, TERA, and VQ-APC, which are significantly smaller than wav2vec2.0 and HuBERT both in model and data size. 
TERA and NPC have comparable model sizes, training objectives, input format, and stride during pre-training, but TERA outperforms NPC with less than one-third of the training data. This is due to the alterations in the time, frequency, and magnitude axes of the data during pre-training, which increases \textbf{data diversity} and enforces accurate phoneme prediction \cite{liu2021tera}. On the other hand, VQ-APC achieves comparable results as NPC with a much smaller model size. With all the other setups identical, this suggests that the \textbf{sequential structure} learned by the Unidirectional LSTM (APC) and the \textbf{quantization layers} are more effective at capturing speech representations than convolutional blocks in NPC, implying that speech should be treated as sequential data.

\subsubsection{Effect of Model Size}\vspace{-0.7mm}
Comparing the HuBERT-\textsc{Base} / HuBERT-\textsc{Large} and wav2vec2.0-\textsc{Base} / wav2vec2.0-\textsc{Large} pairs gives insight into the effect of model size on downstream \ac{asr} tasks. The \textsc{Large} models generally perform better than the \textsc{Base} models.
This is consistent with a previous study by \citet{pu2021scaling}, in which they empirically showed that scaling \ac{ssl} models results in improvements in both L1 loss and accuracy on downstream tasks consistent with the power law. Larger models are also more data-efficient when labeled data is scarce. 
The advantage of the \textsc{Large} model over the \textsc{Base} model is especially apparent on the wav2vec2.0 pair, as wav2vec2.0-\textsc{Large} consistently performs better across all languages. As discussed in Section \ref{sec:results:model:training}, the more efficient use of data in HuBERT-\textsc{Large} may have caused it to learn even more syntactic and semantic representation, which does not benefit cross-lingual speech feature extraction.

\subsection{Linguistic Analysis} \label{sec:results:linguistic}\vspace{-0.7mm}
Now we discuss the performances of all five languages based on their average scores. Smaller $\Delta$ indicates better generalizability. According to the phylogenetic tree shown in Figure \ref{fig:phylotree}, both German and English belong to the Germanic branch; French, Spanish, and Russian are in different language groups as English; Chinese belongs to another language family. As shown in Table \ref{tab:asrresults}, English \ac{ssl} models have better generalizability in French than in German. This is because French has a profound phonological influence on the development of English \cite{roth2010explore}, and the latter not only borrows some French pronunciation rules, but also shares contextual phonetic similarities of pitch contours \cite{so2014phonetic}. For German, although it appears to have poor \ac{ssl} performance with high $\Delta$ values, the absolute WER is the lowest among German, French, and Spanish, which have similar training sizes.
From this, it can be observed that \ac{ssl} representations has diminishing returns in high-resource situations.

Features extracted by the \ac{ssl} models also perform well in Russian and Chinese \ac{asr} tasks. This might seem surprising, but it is because both Russian and Chinese are low-resource with less than 100k utterances. This demonstrates the robustness of \ac{ssl} models in low-resource settings and establishes promising directions to generalize to other low-resource languages. Moreover, although Chinese is in the Sino-Tibetan language family, it actually has some phonotactic similarities with English \cite{ann2014syllabic, yang2021comparison}. It is important to note that the CER was used as the metric for Chinese ASR to avoid additional noise introduced by a word segmentation model, so the Chinese results should only be compared across models rather than across languages.

Analysis by linguistic distance can provide some plausible explanations for the results, but there still exist some inconsistencies. These inconsistencies motivate our next section, PSR Analysis, in which we use our novel metric to explain the model performance by categorizing and quantifying linguistic information in the extracted representations.

\subsection{PSR Analysis}\label{sec:analysis:psr}\vspace{-0.8mm}
\ac{psr} scores of HuBERT-\textsc{Base} on English and the target languages are shown in Table \ref{tab:psr}. As described in Equation \ref{eq:psr}, the larger the \ac{psr}, the more phonetic content in the feature set. 
First, to validate the PSR scale, we test the \ac{ssl} features extracted from an English corpus by the SSL model. The \ac{psr} value from the English corpus is close to zero, which conforms with the intuition that the English-trained HuBERT model is able to extract useful information in both the phonetic and syntactic fields.

\begin{table}[ht]
    \centering\vspace{-0.5mm}
    \scalebox{0.95}{
    \begin{tabular}{cccccc}
    \hline
        Lang & en & de & fr & es & ru\\
    \hline
        \ac{psr} & .01 & .15 & .16&  .13& .23 \\
        WER $\Delta$ & - & 1.3 & 0.7 &  1.6 & -0.1 \\
    \hline
    \end{tabular}
    }\vspace{-2mm}
    \small{\caption{\label{tab:psr}PSR Results for Target Languages. A positive \ac{psr} means that the phonetic content in the extracted representations is stronger than the syntactic content.}\label{tab:allPSR}}\vspace{-2.5mm}
\end{table}

Combined with the information in Table \ref{tab:asrresults}, we show that there is a positive correlation between the \ac{psr} scores of the feature group and the \ac{asr} performance of the model in that language. 
For example, the $\Delta$ value of HuBERT-\textsc{Base} on German is higher (worse) than that of French and lower (better) than that of Spanish as shown in Table \ref{tab:asrresults}, and we see the corresponding relationship of their PSR values in Table \ref{tab:psr}: German PSR is lower (worse, less phonetic info) than French and higher (better, more phonetic info) than Spanish.
This phenomenon indicates that the more phonetic information contained in a set of features, the better the performance of that set of features on cross-lingual or out-of-domain downstream tasks. Therefore, when the \ac{ssl} model trained with English models is applied to the non-English corpus, \textit{phonetic features are the main contributors to effective information compared with syntactic features}.

\subsection{Layer Weights Analysis}\label{sec:results:weights}
\begin{figure}[h]
    \includegraphics[width=0.48\textwidth]{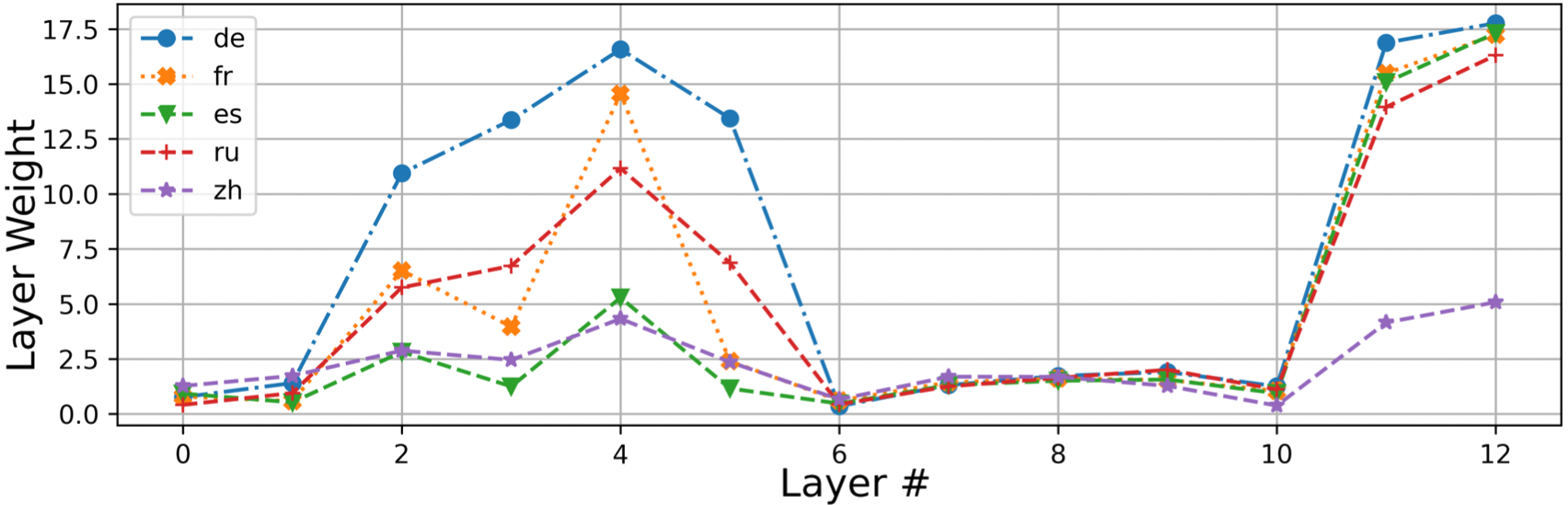}\vspace{-3mm}
    \caption{Layer-wise Weight Analysis.} \label{fig:weights}\vspace{-4.5mm}
\end{figure}

\vspace{-0.8mm}
All PSR scores shown in Table \ref{tab:allPSR} are positive, suggesting that the features extracted by speech \ac{ssl} models tend to have more phonetic information than syntactic information. This is partially due to the fact that the weighted sum of layers is used as input features to the ASR model and that the weights are optimized during training to put more emphasis on the phonetic information. Figure \ref{fig:weights} shows the magnitude of the weights across all layers of HuBERT-\textsc{Base}.


First, the layer-wise trend is consistent across all languages, suggesting each layer contains similar information even when trained on different datasets, i.e., the weights get updated similarly given the same task. 
The optimized weights gravitate toward layers that are crucial for the ASR task. The positive correlation between the ASR and PSR scores implies that the layers with large weights contribute to the high PSR scores, i.e. have denser phonetic than syntactic information.
From Figure \ref{fig:weights}, Layers 4, 11, and 12 contribute significantly to the extracted features. Since lower layers contain lower-level information and vise versa, Layer 4 (and its adjacent layers) contain low or intermediate-level information on acoustic and phonetics important for the \ac{asr} task. The last two layers are the most salient because they contain high-level information related to human phonetics. Additionally, the weight for Layer 4 is larger in German and French, which are closer to English. This shows that when the pre-training and target languages are highly similar, the low-level phonetic features become more helpful. 
Our work to localize the phonetic content encoded in specific layers of HuBERT draws similar conclusions with \citet{pasad2021layer} and \citet{pasad2023comparative}, which localized various acoustic and linguistic properties in \ac{ssl} models using CCA.

\section{Conclusion}\vspace{-1mm}
In this work, we studied English self-supervised speech models and probed for the phonetic and syntactic content in the extracted speech representations. We accomplished this using the \ac{ssl} models as a feature extractor for downstream \ac{asr} task in multiple languages. Higher multilingual adaptability of a model is found to be positively correlated to the amount of phonetic information in the extracted representations. Most importantly, we propose a novel metric - the Phonetic-Syntax Ratio (PSR) - to quantify the phonetic and syntactic composition in the representations. PSR can serve as an effective indicator during model selection. 
We were also able to localize the phonetic information to certain layers in the SSL model.
This is a call to other researchers to design smarter objectives when pre-training large models (such as focusing more on phonetic information learning) rather than simply increasing the model size.

\section*{Limitations}
There are several limitations to our work. First, the value of our PSR was only tested on HuBERT due to limited computing resources. Although the scores reflect the ratio of acoustic and linguistic information in the features extracted by the SSL model, the performance of the corresponding downstream ASR task is not yet empirically shown in every SSL model. Second, the parameters in the SSL models are frozen during \ac{asr} training. Multilingual adaptability might be evaluated differently by unfreezing some or all layers of the SSL feature extractor. Finally, we did not calculate the PSR value for Chinese, as we did not find it to be a valuable data point given the Chinese \ac{asr} results are reported in CER only. Our choice to evaluate English SSL models is motivated by the abundance or English data, but other monolingual or multilingual models could be used given the abundance of data in the chosen langauge(s). For future directions, we believe that exploring spurious correlations among language pairs (e.g. phonotactical similarities between Chinese and English) is a fruitful direction that might shed light on language selection during cross-lingual transfer in speech models.


\bibliography{anthology,custom}
\bibliographystyle{acl_natbib}

\appendix
\section{ASR Model Architecture and Training} \label{app:asr_arch}
The downstream ASR model is composed of a Conformer encoder and a
Transformer decoder. The encoder consists of 12 blocks and 4 attention heads with an output size of 256, and the decoder consists of 6 blocks. We use an Adam optimizer with 25000 warmup steps. The model is initialized with Xavier Uniform distribution and trained for 50 epochs with early stopping. We take the average of the best 10 models as the prediction model in the \ac{asr} task. To focus on the performance of the \ac{ssl} feature extractor, we used a simple stacked RNN as the language model during decoding. The RNN language model has 2 layers and each layer has 650 units optimized by the SGD algorithm. We train this language model for 20 epochs and only keep the best one as our language model. During decoding, we use 0.3 as the weight of the language model and decode data with a beam size of 10.

\section{Implementation and Hardware} \label{app:hardware}
We obtain the upstream \ac{ssl} models and DGCCA model from the S3PRL Speech Toolkit \cite{yang2021superb}.
The \ac{asr} training and DGCCA computation were both done on NVIDIA Tesla V100 for all model-language pairs. The average time of each experiment depends on the dataset size but cost about one week to complete on two GPUs for \ac{asr} and one day for DGCCA.

\end{document}